\documentclass[letterpaper, 10 pt, conference]{ieeeconf}
\IEEEoverridecommandlockouts
\usepackage{cite}
\usepackage{amsmath,amssymb,amsfonts}
\usepackage{algorithm}
\usepackage{algpseudocode}
\usepackage{graphicx}
\usepackage{comment}
\usepackage{url}
\usepackage[colorlinks]{hyperref} 
\usepackage{textcomp}
\usepackage{xcolor}
\usepackage[nolist]{acronym}
\usepackage{tikz}
\usepackage{siunitx}
\usepackage{caption}
\usepackage{xcolor}
\usepackage{subcaption}
\usepackage{booktabs}
\usepackage{multirow}
\usepackage[most]{tcolorbox}
\let\labelindent\relax
\usepackage{enumitem}
\usepackage[utf8]{inputenc} %
\usepackage[T1]{fontenc}    %
\usepackage{colortbl}
\usepackage{color}
\usepackage{setspace}

\newcommand{\x}{\mathbf{x}}

\newcommand{\bbe}{\mathbb{E}}

\RequirePackage{mathtools}
\colorlet{green}{green!30!}
\colorlet{blue}{cyan!30!}
\colorlet{darkgreen}{green!60!black}
\colorlet{darkblue}{cyan!60!}
\colorlet{red}{red!30!}
\colorlet{violet}{violet!30!}
\colorlet{orange}{orange!30!}
\colorlet{cyan}{blue!30!}
\tcbset{
    inline/.style={
        on line, 
        arc=2pt, 
        boxsep=2pt, 
        left=0pt, right=0pt, 
        top=0pt, bottom=0pt, 
        boxrule=0pt, 
        colback=#1, 
        colframe=#1, 
        coltext=black, 
    }
}\newcommand{\mathbox}[2][]{%
  \tcbox[inline=#1]{$#2$}
}

\usetikzlibrary{shapes.geometric, arrows, positioning}

\tikzstyle{block} = [rectangle, rounded corners, minimum width=3cm, minimum height=1cm,text centered, draw=black, fill=blue!20]
\tikzstyle{input} = [ellipse, minimum width=2cm, minimum height=1cm, text centered, draw=black, fill=green!20]
\tikzstyle{output} = [ellipse, minimum width=2cm, minimum height=1cm, text centered, draw=black, fill=red!20]
\tikzstyle{arrow} = [thick,->,>=stealth]

\def\BibTeX{{\rm B\kern-.05em{\sc i\kern-.025em b}\kern-.08em
    T\kern-.1667em\lower.7ex\hbox{E}\kern-.125emX}}

\title{\LARGE \bf Enhancing Physical Consistency in Lightweight World Models}
\author{Dingrui Wang$^{1*}$, Zhexiao Sun$^{1*}$, Zhouheng Li$^{2}$, Cheng Wang$^{4}$, Youlun Peng$^{1}$, Hongyuan Ye$^{1}$,\\
Baha Zarrouki$^{1}$, Wei Li$^{3}$, Mattia Piccinini$^{1}$, Lei Xie$^{2}$, Johannes Betz$^{1}$ \\
\url{https://physics-wm.github.io/}%
\thanks{$^{1}$Professorship of Autonomous Vehicle Systems, TUM School of Engineering and Design, Technical University of Munich, 85748 Garching, Germany; Munich Institute of Robotics and Machine Intelligence (MIRMI), \texttt{\{dingrui.wang, zhexiao.sun, mattia.piccinini, johannes.betz\}@tum.de}}%
\thanks{$^{2}$College of Control Science and Engineering, Zhejiang University, Hangzhou 310027, China. \texttt{zh\_li@zju.edu.cn}}%
\thanks{$^{3}$Nanjing University, China.}
\thanks{$^{4}$School of Mechatronics Engineering, Harbin Institute of Technology, Harbin 150001, China.}%
\thanks{$^{*}$These authors contributed equally to this work. Author order was determined at random.}%
}

\begin{document}

\makeatletter
\let\@oldmaketitle\@maketitle%
\renewcommand{\@maketitle}{\@oldmaketitle%
    \centering
    \vspace{3.8mm}
    \includegraphics[width=1\textwidth]{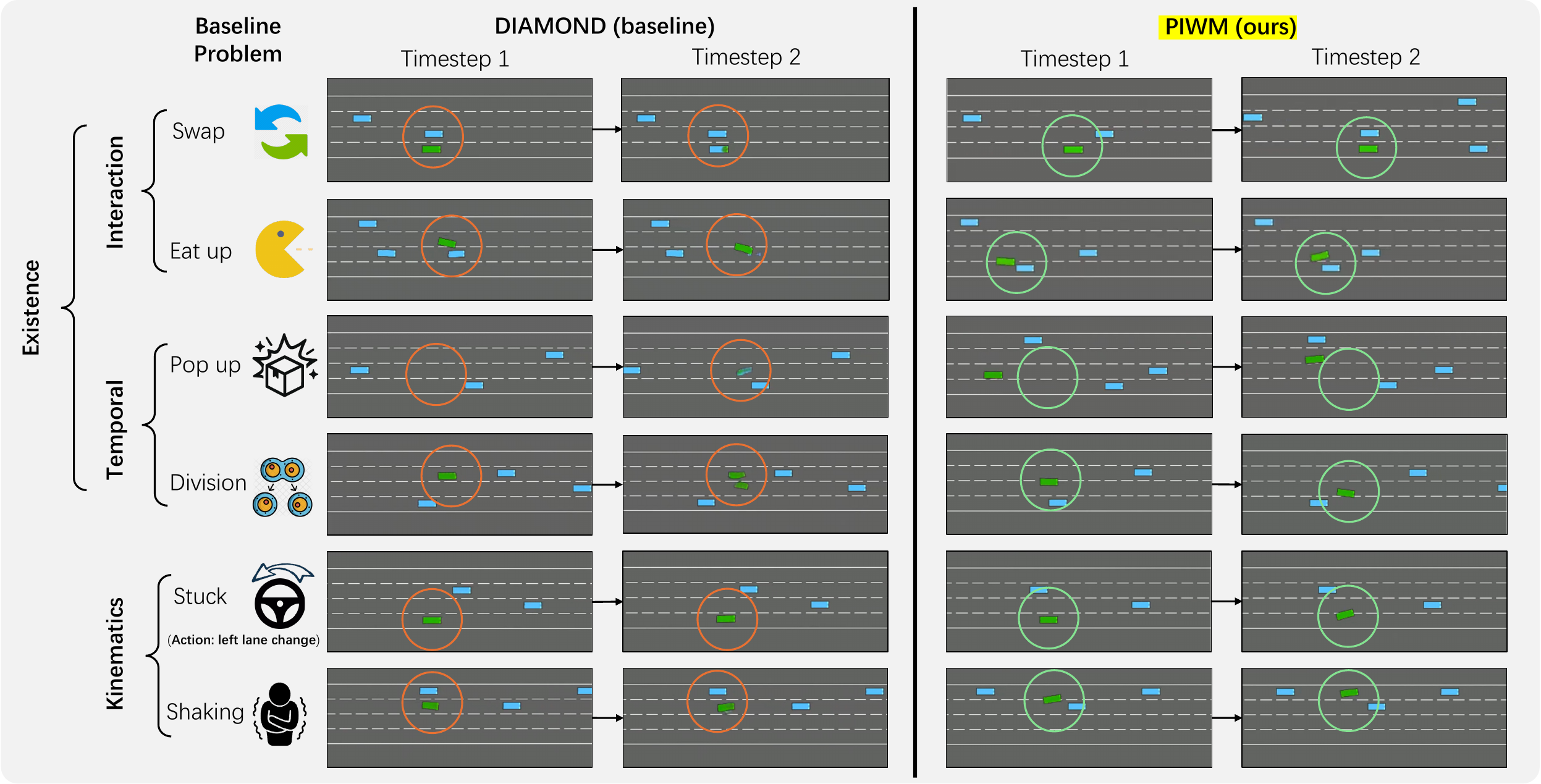}
    \captionof{figure}{\textbf{Performance comparison} between the DIAMOND~\cite{alonso2024diffusion} (baseline) and our Physics-Informed BEV World Model (PIWM), both trained on the dataset collected from HighwayEnv~\cite{highway-env}. PIWM outperforms the baseline by generating more physically consistent results, with better existential consistency and kinematics response.}
    \label{fig:teaser}
    \vspace{-4mm}
}
\makeatother
\maketitle

\setcounter{figure}{1}

\begin{abstract}
A major challenge in deploying world models is the trade-off between size and performance. Large world models can capture rich physical dynamics but require massive computing resources, making them impractical for edge devices. Small world models are easier to deploy but often struggle to learn accurate physics, leading to poor predictions. We propose the Physics-Informed BEV World Model (PIWM), a compact model designed to efficiently capture physical interactions in bird's-eye-view (BEV) representations. PIWM uses \emph{Soft Mask} during training to improve dynamic object modeling and future prediction. We also introduce a simple yet effective techniques—\emph{Warm Start} for inference to enhance prediction quality with zero-shot model. Interactive experiments show that at the same parameter scale (400\text{M}), PIWM surpasses the baseline by 60.6\% in weighted overall score. Moreover, even when compared with the largest baseline model (400\text{M}), the smallest PIWM (130\text{M} Soft Mask) achieves a 7.4\% higher weighted overall score with a 28\% faster inference speed.
\end{abstract}

\section{Introduction}

The Genie series of models~\cite{bruce2024genie, parkerholder2024genie2, genie3} 
has significantly advanced research on world models. Beginning with the original Genie 
and Genie 2, these systems introduced the ability to generate interactive environments 
from data, enabling agents to learn and act within diverse virtual worlds. The most 
recent iteration, Genie 3, demonstrates emergent physical reasoning capabilities: 
it can simulate gravity, collisions, and object interactions without relying on an 
explicit physics engine. This capability indicates that scaling large world models 
can lead to a deeper, implicit understanding of the laws of physics, a critical step 
towards more general embodied intelligence.
Among the world models, bird's-eye-view (BEV) world models~\cite{liao2025diffusiondrive, li2025end} are particularly promising for motion prediction and future trajectory modeling. By representing the environment in a top-down view~\cite{krajewski2018highd, caesar2021nuplan}, BEV models can capture spatial relationships and object interactions more effectively than first-person or ego-centric views, making them suitable for navigation tasks in robotics.

Running world models online is constrained by edge hardware performance. While devices such as the NVIDIA Jetson Orin Nano Super (67 TOPS)\cite{nvidiaJetsonOrinNanoSuper2024}, AGX Orin (200–275 TOPS)\cite{nvidiaJetsonAGXOrin2022}, or Tesla HW4 (500-720 TOPS)\cite{after1989_tesla_hw4} support moderate models, state-of-the-art video/world models like HunyuanVideo\cite{kong2024hunyuanvideo} and Veo 3~\cite{veo3} with hundreds of billions of parameters remain infeasible outside GPU/TPU clusters. More compact models, e.g., DIAMOND (0.4B) runs at 21 FPS on an RTX 4080 Laptop GPU (542 TOPS) but only 9.5 FPS on 100-TOPS edge devices. Both rates fall below the human perceptual smoothness threshold (\(\sim\)24 FPS), which challenges real-time deployment.


While further shrinking the world model to run smoothly on edge devices is feasible, a fundamental challenge remains: can compact models capture the richness of physical dynamics without sacrificing reasoning ability? Reducing model size alone often leads to oversimplification and loss of fidelity. Addressing this gap requires shifting focus from scale to intelligent design—leveraging physics-informed data, inductive biases, and targeted training. Such models can move beyond pattern memorization, enabling robust generalization and effective reasoning under resource constraints.
We address this challenge by extending the 400M-parameter world model DIAMOND~\cite{alonso2024diffusion} to better capture physical consistency (Fig.~\ref{fig:teaser}), by introducing a new Physics-Informed World Model (PIWM) with the following contributions:
%
\begin{itemize}[leftmargin=*]
\item \textbf{Soft Mask.} We propose a method that extracts spatial semantic information, emphasizing the existence of dynamic objects while preserving action sensitivity. Our Soft Mask improves temporal and perceptual video consistency and achieves higher human-judged physics scores than the baseline~\cite{alonso2024diffusion}. Experiments further show that our method enables parameter reduction for edge deployment without compromising physical consistency.

\item \textbf{Warm Starting.} We introduce a zero-shot Warm Start method, that injects contextual information at inference time to improve generation stability at small scales. It can be directly plugged into any pretrained diffusion-based world model, and yields FID gains over the baseline~\cite{alonso2024diffusion}.
\item \textbf{Open-source models and dataset.} We collect 2{,}000 episodes in HighwayEnv~\cite{highway-env} with an MCTS agent, yielding 2 million BEV frames with aligned states and actions. Rewards and actions are carefully designed to evenly cover interaction and lane distributions, thereby supporting future exploration in physical consistency for the community.
\end{itemize}

\section{Related Work}

\noindent \textbf{World Models} have demonstrated the potential to learn complex world dynamics. 
Large-scale models such as the Genie series~\cite{bruce2024genie, parkerholder2024genie2, genie3} and Veo 3~\cite{veo3} show the benefits of scale for long-horizon prediction and richer environment simulation, while recent works~\cite{alonso2024diffusion, bar2025navigation, zhou2024dino} target lightweight models that can run with limited resources. 
DIAMOND~\cite{alonso2024diffusion} trains an RL agent fully inside a 0.4B-parameter diffusion world model.
Inspired by these advances, researchers have applied world models to autonomous driving~\cite{gaia, drivedreamer, li2024enhancing, min2024driveworld}.
Within this domain, Bird's eye view (BEV) has emerged as a promising research direction, leading to attempts of interpreting BEV features as a world model. BEVDiffuser~\cite{bevdiffuser} designs a diffusion model to denoise the BEV feature maps, gaining a significant performance improvement, while Li et al.~\cite{li2025end} introduce WoTE, a BEV world model for efficient and real-time future prediction and trajectory evaluation.


\noindent \textbf{Edge computing.}
Large models are often unsuitable for edge deployment in autonomous driving due to high computational and memory demands. Although compression techniques offer efficiency gains, they pose critical limitations in safety-critical settings. Quantization~\cite{zhang2023qdbev, wang2025ptqat, yu2025qtempfusion} accelerates inference, but amplifies numerical errors under distribution changes. Pruning~\cite{li2024learning, castells2024ldpruner, shirkavand2025efficient} risks removing weights vital for rare scenarios, harming robustness. Knowledge distillation~\cite{yin2024onestep, sauer2024adversarial, nguyen2024swiftbrush} compresses models but often fails to preserve fine-grained spatiotemporal reasoning and precise decision boundaries, degrading performance in tasks. Thus, achieving compact, efficient, and reliable models for real-time edge deployment remains a key challenge. This motivates the design of small models yet capable of accurate and robust dynamic reasoning under resource constraints.

\noindent \textbf{Encoding Physics into World Models.} 
Because large models are hard to deploy on edge devices, distillation~\cite{hinton2015distilling, mishra2017apprentice} is widely used but often yields weaker or unstable performance. We instead train a compact world model tailored for edge constraints.
Beyond efficiency, physics understanding is a key criterion for world models~\cite{duan2025worldscore, worldmodelbench}, indicating whether they capture causal dynamics rather than merely reproducing appearance.
DrivePhysica~\cite{yang2024physical} improves physics awareness in driving world models by enforcing motion in a reference system, temporal consistency, and correct spatial relations.
Using pretrained depth and semantic models, World4Drive~\cite{zheng2025world4drive} can gain a deep understanding of the spatial and semantic properties of the physical world.
Yet maintaining strong physics reasoning in compact models remains open. Here, we show that PIWM effectively strengthens physical understanding under tight resource budgets.

\section{Methodology}
\label{sec:method}

\begin{figure*}[t]
\centering
\includegraphics[width=1.95\columnwidth]{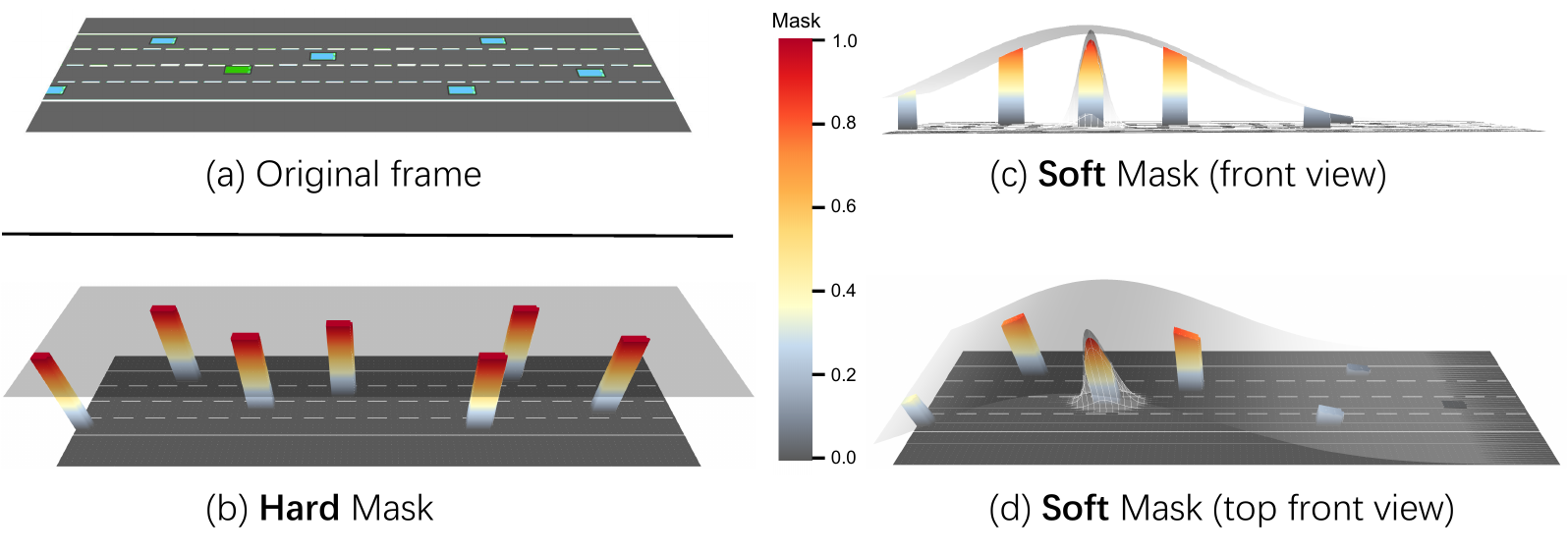}
\vspace{-3mm}
\caption{\textbf{Illustration of soft masks and hard mask.} 
In each 3D cubic shape in the mask figures, the color gradient indicates the mask value according to the chosen mask weights. (a) shows the original frame, while (b) shows the hard mask. (c)-(d) show the Gaussian distributions for global scene softening and ego-centric softening.}
\label{fig:mask_view}
\vspace{-4mm}
\end{figure*}   

This section presents our data collection for world models' training and the design principles of our PIWM. The methods are split into two categories: training stage and inference stage. 
Section \ref{subsec:hardsoft} introduces how we integrate masking into the training stage.
In \ref{subsubsec:training_pipeline}, we elaborate the PIWM's training process, and \ref{subsec:ws} deals with our Warm Start strategy for the inference stage.

\begin{figure}[t]
\centering
\includegraphics[width=1\columnwidth]{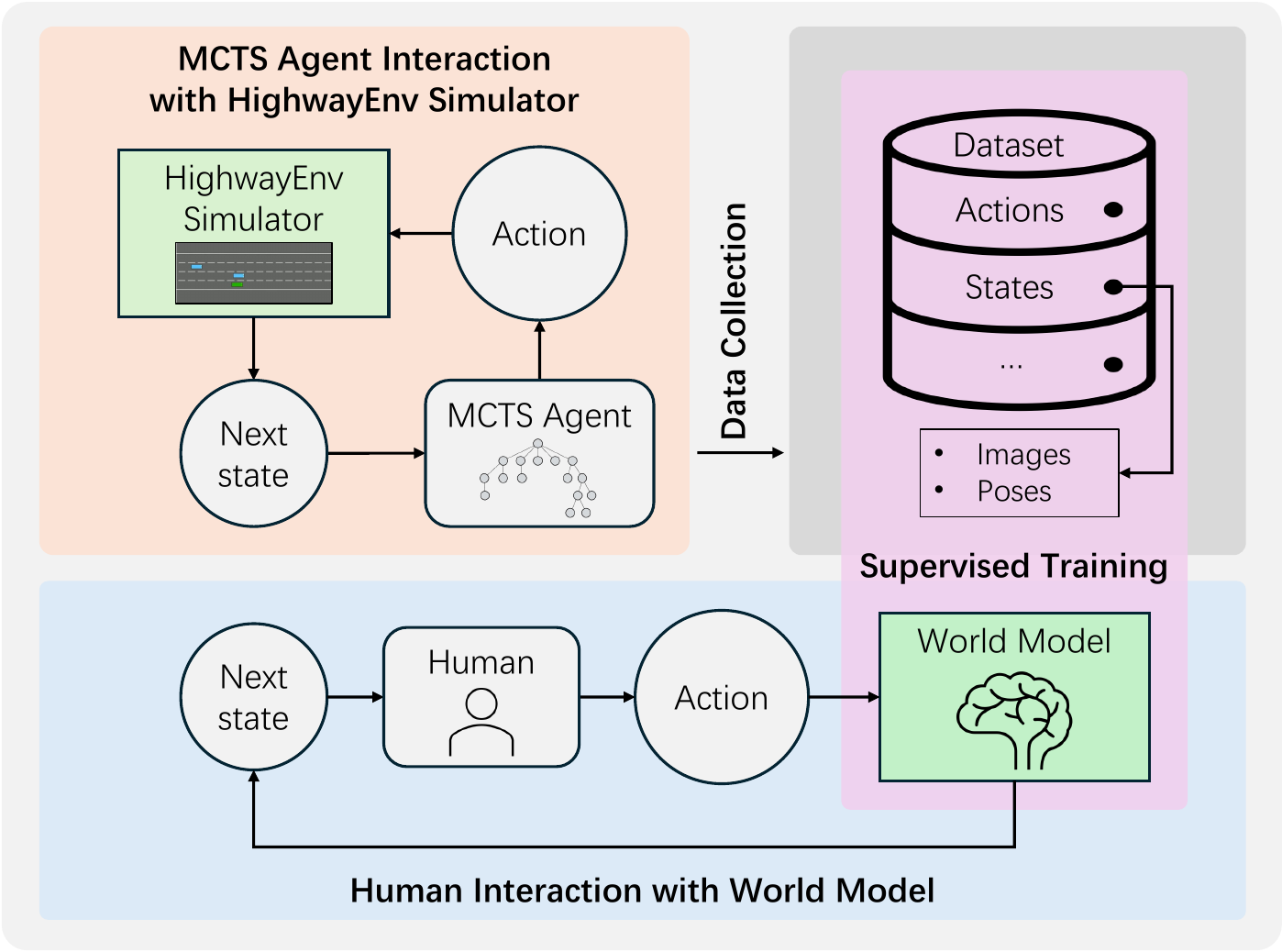}
\caption{The general framework: (Upper Left) Data collection using MCTS agent. (Right) Supervised World Model training. (Bottom) Online inference conditioned on human input.}
\label{fig:framework}
\vspace{-4mm}
\end{figure}   

\subsection{Data Collection}\label{subsec:data_collection}
As illustrated in Fig.~\ref{fig:framework}, we collect training data from the HighwayEnv simulator~\cite{highway-env}, a widely used platform for reinforcement learning with diverse highway scenarios and a BEV representation (Fig.~\ref{fig:teaser}). To promote diversity while maintaining a controlled collision rate, we designed a Monte Carlo Tree Search (MCTS) agent to autonomously control the ego vehicle. The agent is biased toward acceleration maneuvers while maintaining collision avoidance, yielding a broad spectrum of realistic maneuvers. In total, we gather 2{,}000 episodes comprising 2{,}000{,}000 BEV frames, each paired with ground-truth physical states (poses and velocities) for the ego and surrounding vehicles.

\subsection{Training with Hard \& Soft Masks}
\label{subsec:hardsoft}
We use masks to emphasize the existence of dynamic objects and thus aim to enhance physical consistency. Here, the mask can be interpreted as a matrix that shares the same shape as the image and selectively represents its characteristic regions.
To design our masks, we need to first identify the ego vehicle (colored in green in the HighwayEnv simulator, Fig.~\ref{fig:mask_view}a) and the surrounding vehicles (colored in blue) inside a BEV image. 
For this purpose, we adopt a color-checking module which detects the pixels belonging to the ego and the surrounding vehicles.

\subsubsection{Hard Mask}
Following state-of-the-art approaches ~\cite{yang2024physical, wu2025geometry}  that directly leverage geometric information as a condition,
the hard mask employs a \textbf{binary} mask to distinguish between dynamic objects and background environment. Given a BEV image $\mathbf{I} \in \mathbb{R}^{H \times W \times C}$ where $H$ denotes height, $W$ width, and $C$ the number of channels, The hard mask $\mathbf{m}_{\text{hard}} \in \mathbb{N}^{H \times W}$ is constructed as:
\begin{equation}
\mathbf{m}_{\text{hard}}(x, y) = \begin{cases} 
1 & \text{if } (x, y) \text{ is green or blue} \\
0 & \text{otherwise}
\end{cases}
\end{equation}

\subsubsection{Soft Mask}
Inspired by soft constraints in classical control, Soft Mask serves as an input-space soft constraint during both training and inference. It adds a conditioning channel with continuous spatial semantic weights that emphasize interaction-prone dynamic regions while preserving action sensitivity. 
Unlike the hard mask, our soft mask $\mathbf{m}_{\text{soft}} \in [0, 1]^{H \times W}$ assigns continuous values between 0 and 1 to dynamic object regions, while maintaining zero values for static environmental areas. We design $\mathbf{m}_{\text{soft}}$ as:
\begin{equation}
\mathbf{m}_{\text{soft}} =  
(
\color{gray}{\underbracket[1pt]{\color{black}{
w_{\text{ego}} \cdot 
\mathbox[green]{\mathbf{m}_{\text{ego}}} \cdot 
\mathbox[red]{\mathcal{N}_{\text{ego}}}}}_{\mathclap{\textsf{Weighted Gaussian Ego Mask}}}}
+
\color{gray}{\overbracket[1pt]{\color{black}{
w_{\text{surr}} \cdot \mathbox[blue]{\mathbf{m}_{\text{surr}}}}}^{\mathclap{\textsf{Weighted Surrounding Mask}}}}
\color{black}{)
\cdot} 
\mathbox[violet]{\mathcal{N}_{\text{global}}} 
\label{eq_soft_mask}
\end{equation}
In \eqref{eq_soft_mask}, $\mathbf{m}_{\text{ego}}$ and $\mathbf{m}_{\text{surr}}$ are hard masks for the ego vehicle (green channel) and surrounding vehicles (blue channel), weighted by the tunable factors  $w_{\text{ego}}$ and $w_{\text{surr}}$. In practice, we suggest $w_{\text{surr}}$ slightly larger than $w_{\text{ego}}$ during both training and inference to emphasize existential consistency while preserving action sensitivity of the ego vehicle: 
\begin{equation}
\mathbox[green]{
\mathbf{m}_{\text{ego}}}(x, y) = \begin{cases} 
1 & \text{if } (x, y) \text{ is green} \\
0 & \text{otherwise}
\end{cases}
\end{equation}
\begin{equation}
\mathbox[blue]{
\mathbf{m}_{\text{surr}}}(x, y) = \begin{cases} 
1 & \text{if } (x, y) \text{ is blue} \\
0 & \text{otherwise}
\end{cases}
\end{equation}
As shown in Fig.~\ref{fig:mask_view}, we devise two softening dimensions: ego-centric and global scene softening. 
%
\noindent \textbf{Ego-centric Softening} uses a two-dimensional Gaussian distribution centered around the detected ego vehicle:
\begin{equation}
\mathbox[red]{
\mathcal{N}_{\text{ego}}}
=
\mathcal{N}\!\left(
\begin{bmatrix} x \\ y \end{bmatrix}
\;\middle|\;
\begin{bmatrix} x_{\text{ego}} \\ y_{\text{ego}} \end{bmatrix},
\begin{bmatrix}
\sigma_x^2 & 0 \\
0 & \sigma_y^2
\end{bmatrix}
\right)
\end{equation}
where $(x_{\text{ego}}, y_{\text{ego}})$ denote the ego vehicle's centroid, while $\sigma_x$ and $\sigma_y$ are tunable Gaussian parameters.
\textbf{Global Scene Softening} uses a global Gaussian distribution along the longitudinal direction of the BEV scene, centered at the ego vehicle's $x$-coordinate:
\begin{equation}
\mathbox[violet]{
\mathcal{N}_{\text{global}}} = 
\mathcal{N}\!\left(
x \;\middle|\; 
x_{\text{ego}},\,
(\sigma_{\text{global}} \, W)^2
\right)
\end{equation}
where $\sigma_{\text{global}}$ is a tunable parameter regulating the Gaussian's width, and $W$ is the BEV image width.

\noindent \textbf{Downsampling} process uses bicubic interpolation to downsample the softened mask to match the input resolution of the denoising model:
$
\mathbf{m}_{\text{soft}}^{\text{down}} 
\;\leftarrow\; 
\text{bicubic}\!\left(\mathbf{m}_{\text{soft}}\right).
$

\begin{figure}[t]
\centering
\includegraphics[width=1\columnwidth]{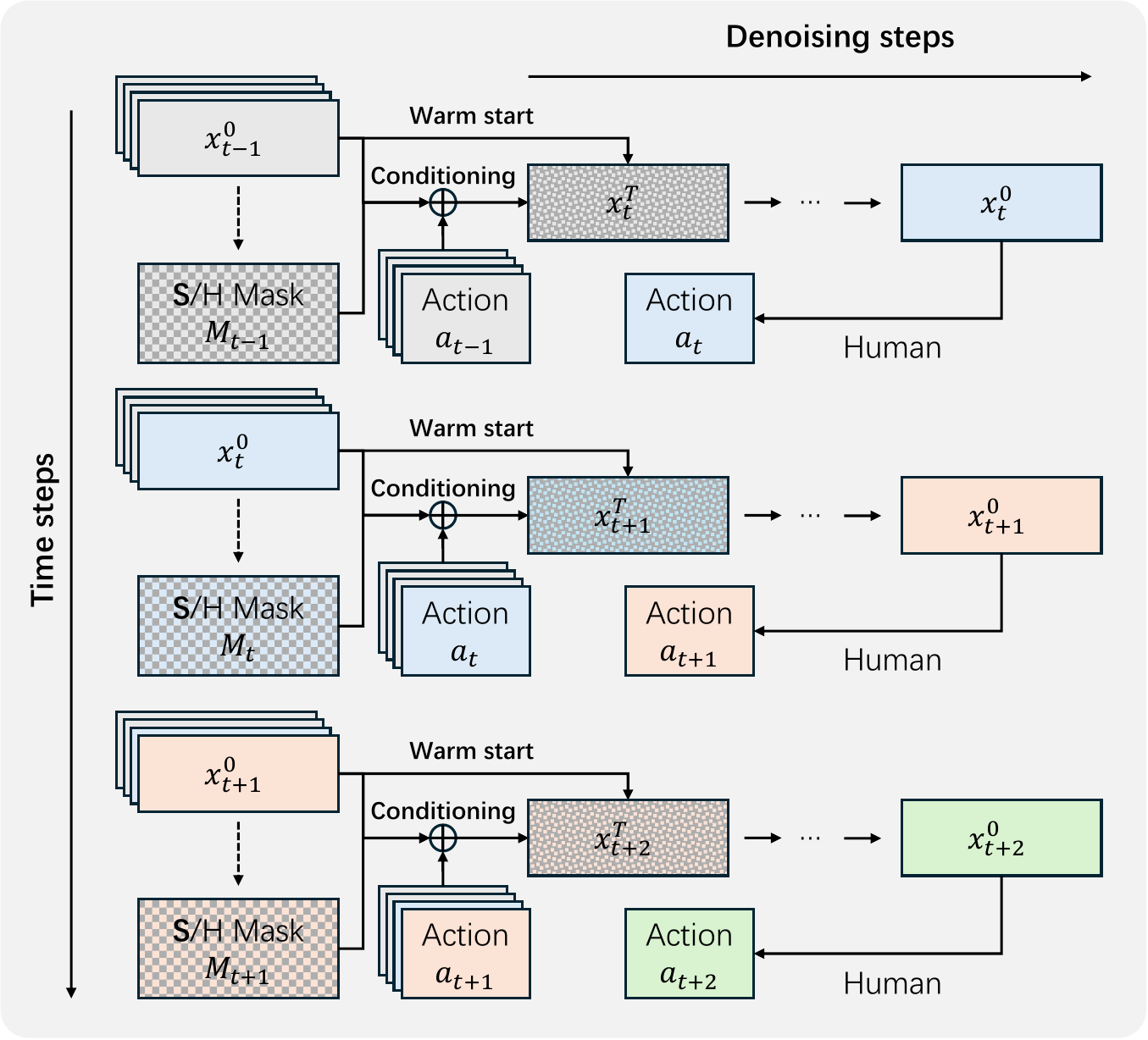}
\caption{Training pipeline.}
\label{fig:training_pipeline}
\vspace{-8mm}
\end{figure}

\subsubsection{World Model Training} \label{subsubsec:training_pipeline}
World Models are generative models that learn to simulate how the environment evolves over time. Given a history of what the agent has seen (observations, actions), the model predicts future observations.
As illustrated in Fig.~\ref{fig:training_pipeline}, we follow a similar EDM~\cite{karras2022elucidating} training setup as the baseline DIAMOND model~\cite{alonso2024diffusion}. Our soft mask is integrated into the EDM architecture as part of the input for the diffusion model $\mathbf{D}_\theta$, where $\theta$ are the trainable parameters. During training, we sample a sequence of length $L$ containing past action–observation pairs, where each pair consists of an action $a$ and an observation $\x$ (image). A sequence at timestamp $t$ is represented as $( \x_{t-L+1}^0, a_{t-L+1}, \dots, \x_t^0, a_t, \x_{t+1}^0 )$, drawn from the dataset $\mathcal{D}$. The denoising process is given as:
\begin{equation}
\label{eq:denoising_sm_conditional}
\mathbox[orange]{\hat{\x}_{t+1}^0} = \mathbf{D}_\theta(\x_{t+1}^\tau, \tau, \x_{t-L+1:t}^0, a_{t-L+1:t}, \mathbf{m}_{\circ, t})
\end{equation}
where $\tau$ is noise level, $\mathbf{m}_{\circ, t}$ can be either $\mathbf{m}_{\text{soft}, t}$ or $\mathbf{m}_{\text{hard}, t}$ ($\mathbf{m}_{\text{soft}, t}$ means Soft Mask at timestep $t$). 
The diffusion model $\mathbf{D}_\theta$ is trained to denoise a corrupted version of $\x_{t+1}^0$, conditioned on the history of observations. To guide the model’s attention, we integrate our soft mask into the EDM architecture by concatenating it with the observations. The training loss is defined as:
\begin{equation}
\label{eq:loss}
     \mathcal{L}(\theta) = \bbe \left[ \Vert \mathbox[orange]{\hat{\x}_{t+1}^0}  - \x_{t+1}^0 \Vert^2 \right].
\end{equation}

\newcommand{\cmt}[1]{\Comment{$\triangleright$ #1}}
\algrenewcommand\algorithmicrequire{\textbf{Require:}}
\algrenewcommand\algorithmicensure{\textbf{Ensure:}}
\begin{algorithm}[t]
  \caption{Physics-Informed BEV World Model}\label{alg:piwm}

  \textbf{Training}\par
  \vspace{-1mm}
 {\setstretch{1.25}
  \begin{algorithmic}[1]
  \Require $\mathcal{D}, P_{mean}, P_{std}$
    \While{not converged}
    \State $ ( \x_{t-L+1}^0, a_{t-L+1}, \dots, \x_t^0, a_t, \x_{t+1}^0 ) \sim \mathcal{D} $ \cmt{sample}
    \State $\log(\sigma) \sim \mathcal{N}(P_{mean}, P_{std}^2)$ \cmt{noise level}
    \State $\tau := \sigma$ \cmt{default identity schedule from EDM}
    \State $\x_{t+1}^\tau \sim \mathcal{N}(\x_{t+1}^0, \sigma^2 \mathbf{I})$ \cmt{add gaussian noise} 
    \State $\mathbox[orange]{\hat{\x}_{t+1}^0} \leftarrow \mathbf{D}_\theta(\x_{t+1}^\tau, \tau, \x_{t-L+1:t}^0, a_{t-L+1:t}, \mathbf{m}_{\circ, t})$
    \State $\mathcal{L} = \Vert \hat{\x}_{t+1}^0 - \x_{t+1}^0 \Vert^2$ \cmt{loss}
    \State $\theta \leftarrow \theta-\nabla_{\!\theta}\mathcal{L}$ \cmt{gradient step}
    \EndWhile
  \end{algorithmic}
 }
  \medskip

  \textbf{Inference}\par
  \vspace{-1mm}
  {\setstretch{1.25}
  \begin{algorithmic}[1]
    \Require noise covariance: $\Sigma_{\mathrm{off}}$ and $\Sigma_{\mathrm{ew}}$.
    \If{$i == 0$}
      \State $\mathbox[cyan]{x_i} \sim \mathcal{N}(0,\sigma_0^2 I)$ \cmt{Initialize image}
    \ElsIf{$i > 0$}
      \State $\mathbox[cyan]{x_i} \sim \mathcal{N}\!\left(x_{i-1,0},\; \Sigma_{\mathrm{off}} + \Sigma_{\mathrm{ew}}\right) $ \cmt{Warm Start} 
    \EndIf
    \State $\tau_0[0] \leftarrow s_0$ \cmt{known initial state}
    \For{$i=0$ to $N-1$}
      \State $\tau_{i+1} \leftarrow S_\theta(\tau_i;\,\sigma_i,\sigma_{i+1})$ \cmt{denoised prediction (2)}
      \State $\tau_{i+1}[0] \leftarrow s_0$ \cmt{known initial state}
    \EndFor
    \State \Return $\tau_{N}$ \cmt{noise-free sample}
  \end{algorithmic}
  }
\end{algorithm}

\begin{figure*}[ht]
\centering
\includegraphics[width=0.95\textwidth]{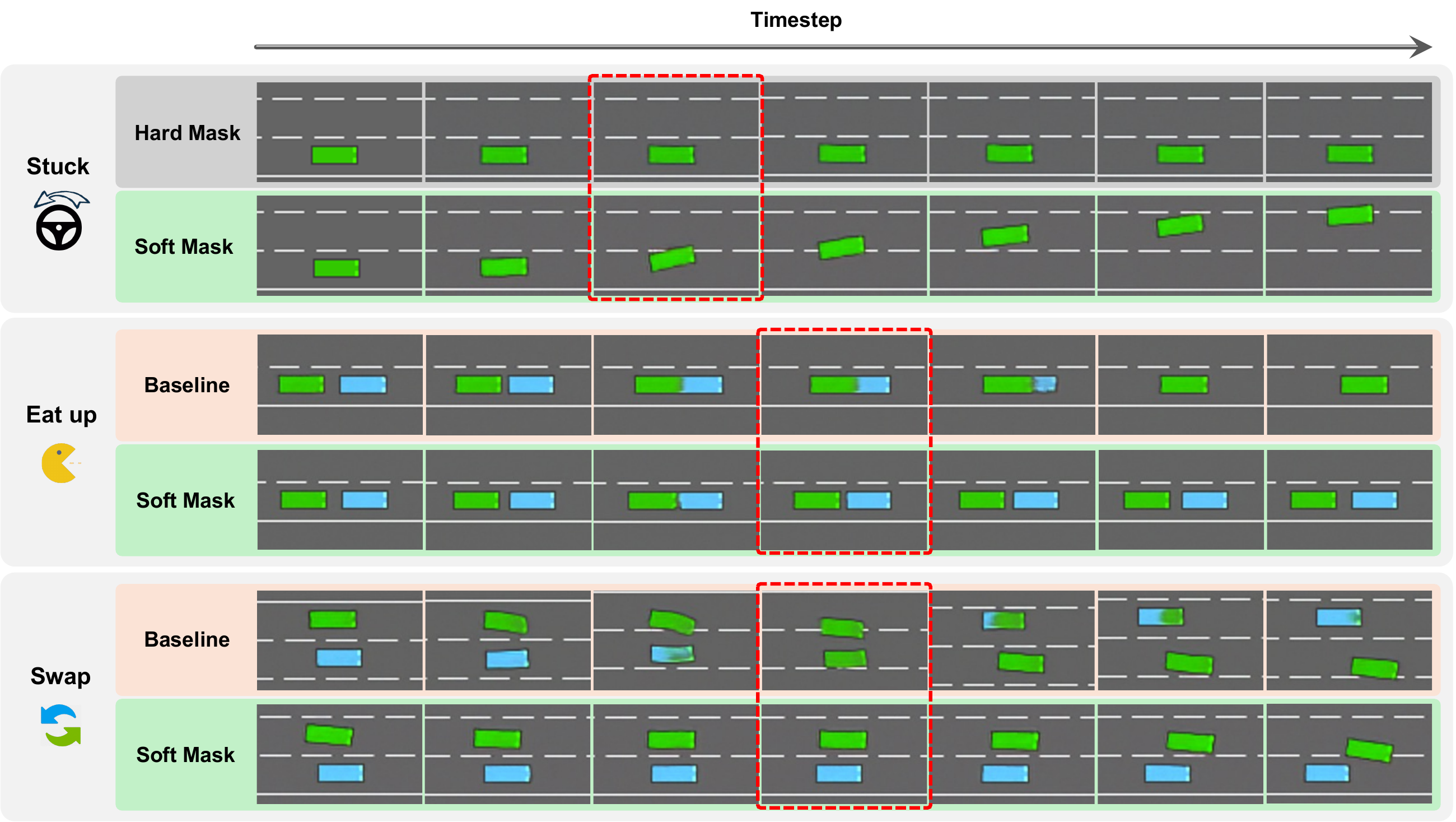}
\caption{\textbf{Qualitative comparison of different methods on common failure cases. }
The first row (“Stuck”) compares Hard Mask and Soft Mask for the “Stuck” issue, which concerns whether the agent is insensitive to actions.
The second row (“Eat up”) and third row (“Swap”) compare the baseline~\cite{alonso2024diffusion} with Soft Mask for the “Eat up” and “Swap” problems, respectively.}
\vspace{-4mm}
\label{fig:qualitative}
\end{figure*}

\subsection{Inference with Warm Starting}
\label{subsec:ws}
To improve temporal consistency in the generative process at inference time, we design a warm-start strategy. The method is model-agnostic (zero-shot) and can be applied to any trained world model without retraining. The intuition follows autoregressive generation: each frame is synthesized by perturbing the previously generated clean frame, which promotes spatial and temporal coherence across the sequence. 
At generation step $i$, instead of sampling from pure Gaussian noise, we initialize the reverse process by perturbing the clean image from step $i-1$, denoted $x_{i-1,0}$. The perturbed sample $\tilde{x}_{i,T}$ is drawn from
\begin{equation}
\begin{aligned}
q(\mathbox[cyan]{\tilde{x}_{i,T}} \mid x_{i-1,0})
&= \mathcal{N}\!\left(x_{i-1,0},\; \Sigma_{\mathrm{off}} + \Sigma_{\mathrm{ew}}\right), \\[6pt]
\Sigma_{\mathrm{off}} &= \sigma_{\mathrm{off}}^{2}\, \mathrm{blkdiag}\!\big(\underbrace{\mathbf{K}_{l}, \dots, \mathbf{K}_{l}}_{C\;\text{times}}\big), \\[-2pt]
\Sigma_{\mathrm{ew}} &= \sigma_{\mathrm{ew}}^{2}\,\mathbf{J}_{n}.
\end{aligned}
\end{equation}
Here, $x_{i-1,0}$ and $\tilde{x}_{i,T}$ are flattened vectors in $\mathbb{R}^{n \times 1}$ with $n=H \times W \times C$; $H$, $W$, and $C$ denote image height, width, and channel count. The matrix $\mathbf{K}_{l}=\mathbf{1}\mathbf{1}^\top \in \mathbb{R}^{l \times l}$ with $l=H \times W$ specifies a rank-1 covariance within each channel, inducing a channel-wise global offset (fully correlated spatial positions). The operator $\mathrm{blkdiag}(\cdot)$ places $C$ copies of $\mathbf{K}_{l}$ along the diagonal, ensuring no cross-channel correlation. The term $\sigma_{\mathrm{ew}}^{2}\mathbf{J}_{n}$ adds element-wise independent noise, where $\mathbf{J}_{n}$ denotes the identity matrix. In practice, $\tilde{x}_{i,T}$ serves as the terminal-time initialization for the denoiser at step $i$, balancing coherence (via $\Sigma_{\mathrm{off}}$) and flexibility for local changes (via $\Sigma_{\mathrm{ew}}$).



\section{Results}


\subsection{Experimental Setup}

We evaluate action-conditioned BEV highway video generation: given the last four frames and the current action, the model predicts the next frame and then rolls out autoregressively.
We compare four variants using the same U\!-Net backbone under the EDM framework: (i) \textbf{Baseline} (DIAMOND~\cite{alonso2024diffusion}), (ii) \textbf{Hard Mask}, (iii) \textbf{Warm Start}, and (iv) \textbf{Soft Mask}. Method-specific mechanisms were detailed in~\ref{subsec:hardsoft} and~\ref{subsec:ws}. We report three parameter budgets (130\text{M}, 170\text{M}, 400\text{M}) by changing only the denoiser channel width. Consistent with our evaluation scope, reconstruction quality was reported for baseline/Warm Start/Soft Mask; physical consistency was reported for all four methods; efficiency was reported for baseline/Soft Mask. 
\subsubsection{Reconstruction Quality} 
We sample 100 initial 16-frame segments from the test set as real observations and generate corresponding 16-frame rollouts with identical spawn points (Starting player position) and action sequences. FID is computed with \texttt{pytorch-fid} (Inception-V3 (\texttt{pool3})) over all frames jointly. FVD is computed with \texttt{cd-fvd} (I3D backbone). LPIPS loss is computed with \texttt{lpips} (AlexNet backbone) between paired real/generated frames and averaged over spawn points.

\begin{table}[ht]
\begin{center}
\setlength{\tabcolsep}{8pt}
\renewcommand{\arraystretch}{1.2}
\small
\begin{tabular}{cc|ccc}
\toprule
\multirow{2}{*}{Parameters} & \multirow{2}{*}{Method} & \multicolumn{3}{c}{Reconstruction metrics} \\ 
&  & FID~$\downarrow$ & FVD~$\downarrow$ & LPIPS~$\downarrow$\\
\midrule
\multirow{3}{*}{130M} & Baseline & 52.9 & 304.1 & 0.021 \\
                        & Warm Start & \textbf{50.9} & 298.9 & 0.022  \\
                        & Soft Mask & 74.1 & \textbf{269.4} & 0.023 \\ \midrule
\multirow{3}{*}{170M} & Baseline & 36.4 & 362.8 & 0.022 \\
                        & Warm Start & \textbf{33.3} & 367.8 & 0.022 \\ 
                        & Soft Mask & 79.7 & \textbf{189.3} &0.026 \\ \midrule
\multirow{3}{*}{400M} & Baseline & 23.3 & 204.2 & 0.013 \\
                        & Warm Start & \textbf{22.2} & 209.4 & 0.013 \\ 
                        & Soft Mask & 52.1 & \textbf{156.8} & 0.014 \\ 
                        \bottomrule
\end{tabular}
\end{center}
\vspace{-2mm}
\caption{Results of FID, FVD and LPIPS.}
\label{tab:reconstruction_metric}
\vspace{-4mm}
\end{table}

\subsubsection{Physics Consistency} 
As noted by the creators of Genie~3, evaluating visual world models can be subjective.\footnote{\href{https://youtu.be/ekgvWeHidJs?si=rD5NK3cFa5MRJ1EV&t=1525}{\textit{How Do You Measure the Quality of a World Model?}}}
To obtain quantitative judgments, we adopt Mean Opinion Score (MOS) as our primary evaluation method, following ITU-R BT.500~\cite{MOS1,MOS2} on a five-point scale (10=Excellent, 8=Good, 6=Fair, 4=Poor, 2=Bad). We conduct a within-subject, double-blind user study with 24 non-expert raters, who evaluated all four methods under the same user interface and task prompts. We report Interactive Existential Consistency (IEC), Kinematics Response (KIR), and Temporal Existential Consistency (TEC), averaged and mapped to a percentage scale.
. We also report a weighted overall score (WO) emphasizing interactive existential stability:
\begin{equation}
  \text{WO} = 0.5 \cdot \text{IEC} + 0.25 \cdot \text{KIR} + 0.25 \cdot \text{TEC}.
\end{equation}
\subsubsection{Efficiency}
We benchmark on three platforms: RTX~4090, RTX~4080 (laptop), and RTX~3060 (laptop). For each parameter budget and configuration, we run 10 trials of 1{,}000 frames under identical resolution, sampling settings, initial frames, and action sequence; the first five frames of each run are discarded. We report p95 FPS, p95 inference-only latency measured with CUDA events, and peak GPU memory, averaged across trials.


\vspace{-2mm}

\begin{figure}[t]
    \centering
    \begin{subfigure}[b]{0.15\textwidth}
        \centering
        \includegraphics[width=\textwidth]{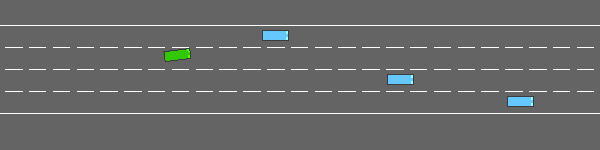}
        \caption{Ground truth.}
        \label{fig:short_gt}
    \end{subfigure}
    \begin{subfigure}[b]{0.15\textwidth}
        \centering
        \includegraphics[width=\textwidth]{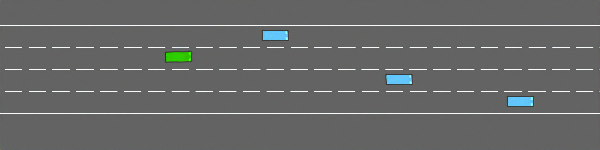}
        \caption{Baseline.}
        \label{fig:short_bl}
    \end{subfigure}
    \begin{subfigure}[b]{0.15\textwidth}
        \centering
        \includegraphics[width=\textwidth]{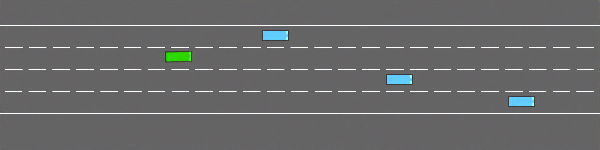}
        \caption{Soft Mask.}
        \label{fig:short_soft}
    \end{subfigure}
    \begin{subfigure}[b]{0.167\textwidth}
        \centering
        \includegraphics[width=\textwidth]{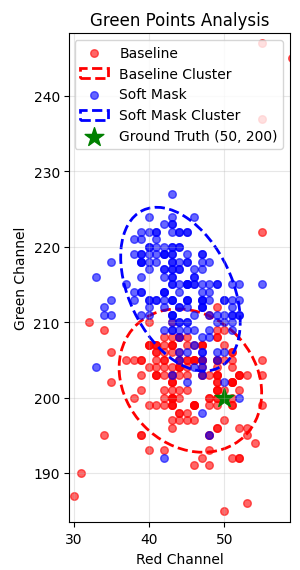}
        \caption{Green Points.}
        \label{fig:green_pts}
    \end{subfigure}
    \begin{subfigure}[b]{0.3118\textwidth}
        \centering
        \includegraphics[width=\textwidth]{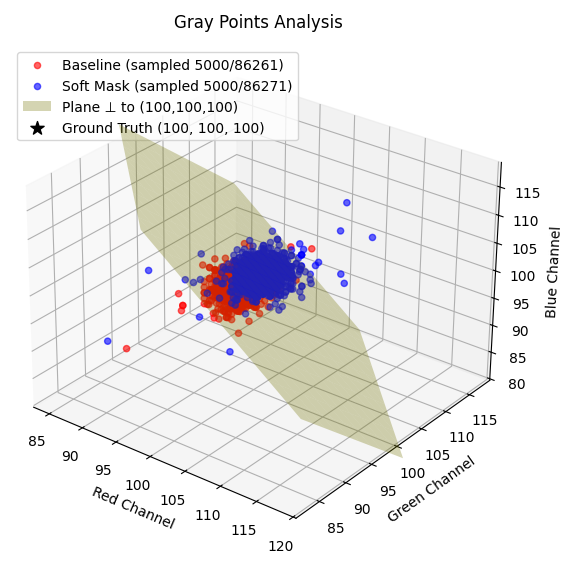}
        \caption{Gray Points.}
        \label{fig:gray_pts}
    \end{subfigure}
    \caption{Color distribution difference concerning the RGB channel values for baseline and Soft Mask.}
    \label{fig:color_diff}
    \vspace{-6mm}
\end{figure}

\subsection{Reconstruction Quality}
At three model sizes (130\text{M}, 170\text{M}, 400\text{M}), Table~\ref{tab:reconstruction_metric} shows three consistent trends. 
\textbf{(i)} Soft Mask achieves the best FVD at all scales—$269.4$ (130\text{M}), $189.3$ (170\text{M}), $156.8$ (400\text{M})—improving over 
baseline by $11.4\%$, $47.8\%$, and $23.2\%$, respectively, indicating stronger temporal coherence.
\textbf{(ii)} Warm Start yields a small but consistent FID gain over baseline.
\textbf{(iii)} LPIPS is low for all methods (all $<0.03$), which reflecting high perceptual similarity to ground truth and differences between methods are minor.

Visual and distributional analysis (Fig.~\ref{fig:color_diff}) suggests the higher FID for Soft Mask is driven by color-distribution shifts rather than perceptible degradations in image quality: compare to baseline, Soft Mask slightly increases green-channel intensity on green pixels and RGB values on gray background (Fig.~\ref{fig:green_pts}, \ref{fig:gray_pts}), which changes dataset-level statistics that FID captures despite minimal visual differences between Fig.~\ref{fig:short_bl} and Fig.~\ref{fig:short_soft}.

In summary, perceptual quality is comparable across methods (low LPIPS), while FVD—more sensitive to motion quality—consistently favors Soft Mask at all parameter sizes, indicating better temporal dynamics.

\begin{table}[t]
\begin{center}
\setlength{\tabcolsep}{7pt}
\renewcommand{\arraystretch}{1.2}
\small
\begin{tabular}
{cccccc}
\toprule
Param. & \multicolumn{1}{l|}{Method} & IEC $\uparrow$ & \multicolumn{1}{c}{KIR $\uparrow$} & \multicolumn{1}{c|}{TEC $\uparrow$} & \multicolumn{1}{c}{WO $\uparrow$}  \\ \midrule
\multirow{2}{*}{75M} 
& \multicolumn{1}{l|}{Baseline$^\dagger$} & 22.50 & 45.00 & \multicolumn{1}{c|}{47.50} & \multicolumn{1}{c}{34.38}  \\ 
& \multicolumn{1}{l|}{Warm Start$^\dagger$} & 22.50 & 50.00 & \multicolumn{1}{c|}{52.50} & \multicolumn{1}{c}{\textbf{36.88}}  \\ 
 \midrule
\multirow{3}{*}{130M} 
& \multicolumn{1}{l|}{Baseline} & 28.12 & 53.59 & \multicolumn{1}{c|}{56.46} & \multicolumn{1}{c}{41.57}  \\ 
& \multicolumn{1}{l|}{Hard Mask$^\dagger$} & 38.75 & 32.50 & \multicolumn{1}{c|}{17.50} & \multicolumn{1}{c}{31.88}  \\
& \multicolumn{1}{l|}{Soft Mask} & \textbf{43.75} & \textbf{64.17} & \multicolumn{1}{c|}{\textbf{56.87}} & \multicolumn{1}{c}{\textbf{52.14}}  \\
 \midrule
 \multirow{3}{*}{170M} 
& \multicolumn{1}{l|}{Baseline} & 32.09 & 51.04 & \multicolumn{1}{c|}{69.38} & \multicolumn{1}{c}{46.15}  \\ 
& \multicolumn{1}{l|}{Hard Mask$^\dagger$} & 42.50 & 46.25 & \multicolumn{1}{c|}{27.50} & \multicolumn{1}{c}{39.69}  \\
& \multicolumn{1}{l|}{Soft Mask} & \textbf{60.21} & \textbf{63.16} & \multicolumn{1}{c|}{\textbf{75.83}} & \multicolumn{1}{c}{\textbf{64.85}}  \\
 \midrule
\multirow{3}{*}{400M}  
& \multicolumn{1}{l|}{Baseline} & 30.63 & \textbf{70.63} & \multicolumn{1}{c|}{62.29} & \multicolumn{1}{c}{48.55} \\ 
& \multicolumn{1}{l|}{Hard Mask$^\dagger$} & 47.50 & 38.75 & \multicolumn{1}{c|}{40.00} & \multicolumn{1}{c}{43.44}  \\
& \multicolumn{1}{l|}{Soft Mask} & \textbf{82.08}& 64.68 & \multicolumn{1}{c|}{\textbf{82.90}} & \multicolumn{1}{c}{\textbf{77.94}} \\

\bottomrule
\end{tabular}
\end{center}
\vspace{-0.6em}
\caption{Results of Human evaluation scores. The metrics considered are Interactive Existential Consistency (IEC), Kinematics Response (KIR), and Temporal Existential Consistency (TEC), Weighted Overall (WO). Baseline is DIAMOND~\cite{alonso2024diffusion}. $^\dagger$ indicate the experiments are evaluated by 4 humans. While the rest are evaluated by 24 humans.
}
\label{tab:sota}
\vspace{-6mm}
\end{table}

\begin{table*}[t]
\begin{center}
\setlength{\tabcolsep}{9pt}
\renewcommand{\arraystretch}{1.2} 
\small
\begin{tabular}{cc|p{0.7cm}cp{1cm}|p{0.7cm}cp{1cm}|p{0.7cm}cp{1cm}}
\toprule
\multirow{3}{*}{Param.} & \multirow{3}{*}{Method} 
& \multicolumn{3}{c|}{RTX 4090} 
& \multicolumn{3}{c|}{RTX 4080 Laptop} 
& \multicolumn{3}{c}{RTX 3060 Laptop} \\ 

& & \multicolumn{3}{c|}{\small (1321 TOPS)} 
  & \multicolumn{3}{c|}{\small (542 TOPS)} 
  & \multicolumn{3}{c}{\small (105 TOPS)} \\ 

& & FPS~$\uparrow$ & Latency~↓ & GPU~↓
  & FPS~$\uparrow$ & Latency~↓ & GPU~↓ 
  & FPS~$\uparrow$ & Latency~↓ & GPU~↓ \\
\midrule
\multirow{2}{*}{130\text{M}} & Baseline & 32.61 & 16.93 & 834.3
                      & 28.14 & 29.98 & 749.6 
                      &  12.41 & 71.90 & 923.4 \\
                       & Soft Mask & 32.19 & 17.48 & 834.5 
                       & 27.99 & 30.39 & 749.7
                       & 12.40 & 72.04 & 923.6 \\ 
\midrule
\multirow{2}{*}{170\text{M}} & Baseline & 32.16 & 17.15 & 975.8 
                        & 27.40 & 31.23 & 889.0
                        & 12.12 & 73.88 & 1064.8 \\
                        & Soft Mask & 30.99 & 17.34 & 974.7
                        & 27.53 & 30.89 & 889.8 
                        &  12.04 & 74.42 & 1063.7 \\ 
\midrule
\multirow{2}{*}{400\text{M}} & Baseline & 27.49 & 21.69 & 1795.7 
                        & 21.83 & 40.17 & 1727.9
                        & 9.72 & 93.42 & 1884.8 \\
                        & Soft Mask & 28.32 & 22.08 & 1795.9 
                        & 21.37 & 41.82 & 1727.0 
                        &  9.67 & 93.99 & 1885.0 \\
\bottomrule
\end{tabular}
\end{center}
\vspace{-2mm}
\caption{Results of FPS, latency, and peak GPU memory usage. GPU memory in MiB. FPS indicates 95th percentile (p95) Frames per second. Latency in ms. All measurements with compilation enabled.}
\label{tab:efficiency_1}
\vspace{-6mm}
\end{table*}

\subsection{Physical Consistency}
\label{subsec:physics}
Beyond generative quality, adherence to physical constraints more directly determines practical utility. We therefore conduct qualitative and quantitative evaluations. The first row ("Stuck") in Fig.~\ref{fig:qualitative} shows that the binary spatial guidance of Hard Mask can over-constrain behavior, suppressing actions and even preventing lane changes. Under the same scenario and action inputs, Soft Mask applies weighted, continuous spatial-semantic guidance that preserves object existence while remaining action-sensitive. The second ("Eat up") and third ("Swap") rows illustrate that the baseline frequently exhibits existential errors in interactive scenes, while Soft Mask is better in object continuity.

Quantitatively, we evaluated all methods using a double-blind, within-subject MOS protocol; results are reported in Table~\ref{tab:sota}. Hard Mask increases IEC relative to baseline (e.g., +37.8\% @130\text{M}, +55.1\% @400\text{M}) but, due to over-constraint, degrades KIR and TEC (e.g., -39.3\% / -69.2\% @130\text{M}), consistent with the "Stuck" failure mode in Fig.~\ref{fig:qualitative}. Warm Start, as a training-free zero-shot approach, yields modest gains at small model scales. At 75\text{M} with four expert raters, it shows little change in IEC but improves kinematic stability (KIR +11.1\%) and temporal stability (TEC +10.5\%), producing a WO gain of +7.3\% without retraining.

A detailed per-metric comparison between Soft Mask and baseline is shown below.
\subsubsection{Interactive Existential Consistency (IEC)}
IEC quantifies identity/existence preservation under interactive, extreme maneuvers (e.g., collisions, squeezes). Soft Mask significantly outperforms baseline at all scales: \textbf{+55.6\%} (130\text{M}), \textbf{+87.6\%} (170\text{M}), \textbf{+168.1}\% (400\text{M}). Baseline’s IEC scores average around 30 with limited gains from scaling, whereas Soft Mask’s continuous pixel-weighted guidance stabilizes identity and existence, achieving \textbf{82.08} at 400\text{M}.

\subsubsection{Kinematics Response (KIR)}
KIR captures immediate controllability and the amplitude response to actions. Soft Mask maintains an advantage at small-to-medium scales: +19.7\% (130\text{M}) and +23.8\% (170\text{M}); at 400\text{M} it is slightly below baseline (-8.4\%). Together with improved FVD at each scale, the results suggest comparable or better kinematic behavior via smoothed per-step action amplitudes.

\subsubsection{Temporal Existential Consistency (TEC)}
TEC measures existence consistency over time. Soft Mask leads by +9.3\% and +33.1\% at 170\text{M}  and 400\text{M}, respectively, and is on par with baseline at 130\text{M}, indicating that continuous spatial modulation improves cross-timestep stability. 

Across metrics, our proposed Soft Mask dominates IEC and TEC while remaining competitive on KIR, yielding markedly higher weighted overall (WO) than baseline: \textbf{+25.4\%} (130\text{M}), \textbf{+40.5\%} (170\text{M}) , and \textbf{+60.6\%} (400\text{M}). Notably, even the smallest Soft Mask model (130\text{M}) achieves higher IEC and WO than the largest baseline model (400\text{M}). Overall, continuous spatial-semantic guidance provided by Soft Mask mitigates the "stuck" issue of Hard Mask and significantly improves interactive and long-term physical consistency without sacrificing action flexibility.

\subsection{Edge Computing Efficiency}
As shown in Table~\ref{tab:efficiency_1}, we compare baseline and Soft Mask across parameter scales and hardware spanning a broad TOPS (Tera Operations Per Second) range representative of robotics and autonomous driving deployments. Soft Mask matches baseline’s resource footprint across settings, adding no measurable computational overhead. On a representative $542$~TOPS device, downsizing to 170\text{M} reduces p95 latency by \(\sim\)25\% and increases p95 FPS from \(\sim\)21 to \(\sim\)27, surpassing the 24~FPS perceptual smoothness threshold. Further downsizing to 130\text{M}, and combined with Table~\ref{tab:sota}, the 130\text{M} Soft Mask attains a WO of \textbf{52.14} at \textbf{27.99} FPS, whereas the 400\text{M} baseline attains \textbf{48.55} at \textbf{21.83} FPS (below 24~FPS), supporting parameter reduction as a practical path for edge deployment without compromising physical consistency.

\section{Discussion}

We first establish a clear efficacy anchor at a higher parameter budget (400\text{M}) and then verify scale-insensitive improvements at smaller budgets (170\text{M} and 130\text{M}). Under edge computing, when p95 FPS meets real-time display (e.g., 24~FPS), smaller configurations still outperform the larger baseline on physics-oriented human scores. This indicates that the gains stem from the mechanisms themselves rather than from parameter count, and it supports parameter reduction as a practical path to edge deployment without sacrificing physics consistency. Trends on TEC further suggest that continuous spatial modulation improves cross-timestep stability, consistent with the intended role of Soft Mask in stabilizing temporal dynamics.

FID results reveal that Soft Mask may exhibit channel-level shifts during generation, likely caused by out-of-distribution deviations in generalization. Nevertheless, this does not compromise practical utility: considering perceptual similarity (LPIPS) and temporal consistency (FVD and TEC), Soft Mask consistently achieves superior performance.

Limitations remain. First, our evaluation is conducted solely in simulation using HighwayEnv, which may not capture the full complexity of real-world driving scenarios including adverse weather conditions, complex urban environments, etc. Second, the current mask construction relies on color-based detection and could be made more robust. Finally, the physics consistency evaluation relies primarily on subjective Mean Opinion Scores (MOS), which, despite following established standards, may miss subtle physical violations critical in safety-critical applications.



\section{Conclusion and Future Work}
We present a Physics-Informed BEV World Model (PIWM) that improves physical consistency while avoiding action suppression and additional computational overhead. PIWM provides two mechanisms: Soft Mask, a training-time conditioning channel with continuous spatial semantic weights that highlight interaction-prone regions while preserving action sensitivity, and Warm Start, a training-free inference strategy that enhances generation stability. Across parameter scales, PIWM with Soft Mask substantially improves physics-oriented human scores and generative dynamics metric (FVD), and under edge-computing budgets that satisfy real-time display, even smaller models can maintain better physical consistency than larger baseline.

\textbf{Future work} will adapt PIWM to real-world datasets with richer conditions and rare events, develop a decoder that directly predicts future states to guide generation and reduce drift, and explore general objective physics metrics for comprehensive evaluation of spatiotemporal consistency and dynamics. We also plan to evaluate our methods in closed-loop planning and control to assess safety and robustness under domain shifts.

\vspace{-2mm}

\bibliographystyle{IEEEtran}
\bibliography{literature.bib}

\begin{acronym}
\acro{AVs}{autonomous vehicles}
\acro{RL}{reinforcement learning}
\end{acronym}

\end{document}